\documentclass{llncs}

\usepackage{url}
\usepackage{amsmath}
\usepackage{todonotes}
\usepackage{amsfonts}
\usepackage{tabularx}
\usepackage{multirow}

\begin{document}

\title{Generalising Random Forest Parameter Optimisation to Include Stability and Cost}
\titlerunning{Generalising Parameter Optimisation}  % abbreviated title (for running head)
%                                     also used for the TOC unless
%                                     \toctitle is used
%

\author{C.H. Bryan Liu\inst{1} \and Benjamin Paul Chamberlain\inst{2} \and Duncan A. Little\inst{1} \and \^{A}ngelo Cardoso\inst{1}}
\authorrunning{Liu et al.} % abbreviated author list (for running head)
%
%%%% list of authors for the TOC (use if author list has to be modified)
\tocauthor{C.H. Bryan Liu, Benjamin Paul Chamberlain,
Duncan Little and \^{A}ngelo Cardoso}
\institute{ASOS.com, London, UK\\
\email{bryan.liu (at) asos.com}
\and
Department of Computing, Imperial College London, London, UK}

\maketitle              % typeset the title of the contribution

\begin{abstract}

% Describe the task/problem the paper is going to address (high level)
% Applications of random forests in industry and academia are becoming increasingly popular.
% Why is this an interesting/important problem?
Random forests are among the most popular classification and regression methods used in industrial applications. To be effective, the parameters of random forests must be carefully tuned. This is usually done by choosing values that minimize the prediction error on a held out dataset. We argue that error reduction is only one of several metrics that must be considered when optimizing random forest parameters for commercial applications. We propose a novel metric that captures the stability of random forest predictions, which we argue is key for scenarios that require successive predictions. 
% How does one usually solve this?
We motivate the need for multi-criteria optimization by showing that in practical applications, simply choosing the parameters that lead to the lowest error can introduce unnecessary costs and produce predictions that are not stable across independent runs.
To optimize this multi-criteria trade-off, we present a new framework that efficiently finds a principled balance between these three considerations using Bayesian optimisation. The pitfalls of optimising forest parameters purely for error reduction are demonstrated using two publicly available real world datasets.
% Interpretation of the results (impact and importance)
We show that our framework leads to parameter settings that are markedly different from the values discovered by error reduction metrics alone.

\keywords{Bayesian optimisation; parameter tuning; random forest; machine learning application; model stability}
\end{abstract}

\section{Introduction}
\label{sec:intro}

% what is a RF
Random forests are ensembles of decision trees that can be used to solve classification and regression problems. They are very popular for practical applications because they can be trained in parallel, easily consume heterogeneous data types and achieve state of the art predictive performance for many tasks \cite{Fernandez-Delgado2014,Tamaddoni2016,Vanderveld2016}. 

% random forest parameters
Forests have a large number of parameters (see \cite{Criminisi2011}) and to be effective their values must be carefully selected \cite{Huang2016}. This is normally done by running an optimisation procedure that selects parameters that minimize a measure of prediction error. A large number of error metrics are used depending on the problem specifics. These include prediction accuracy and area under the receiver operating characteristic curve (AUC) for classification, and mean absolute error (MAE) and root mean squared error (RMSE) for regression problems. Parameters of random forests (and other machine learning methods) are optimized exclusively to minimize error metrics. We make the case to consider monetary cost in practical scenarios and introduce a novel metric which measures the stability of the model.

% the random in RF
Unlike many other machine learning methods (SVMs, linear regression, decision trees), predictions made by random forests are not deterministic. While a deterministic training method has no variability when trained on the same training set, it exhibits randomness from sampling the training set. We call the variability in predictions due solely to the training procedure (including training data sampling) the \textbf{endogenous variability}.
% instability related work
It has been known for many years that instability plays an important role in evaluating the performance of machine learning models. The notion of instability for bagging models (like random forests) was originally developed by Breiman \cite{Breiman1996a,Breiman1996}, and extended explicitly by Elisseeff et. al. \cite{Elisseeff2005} to randomised learning algorithms, albeit focusing on generalisation/leave-one-out error (as is common in computational learning theory) rather than the instability of the predictions themselves.

\begin{figure}
    \begin{center}   \includegraphics[width=0.9\textwidth]{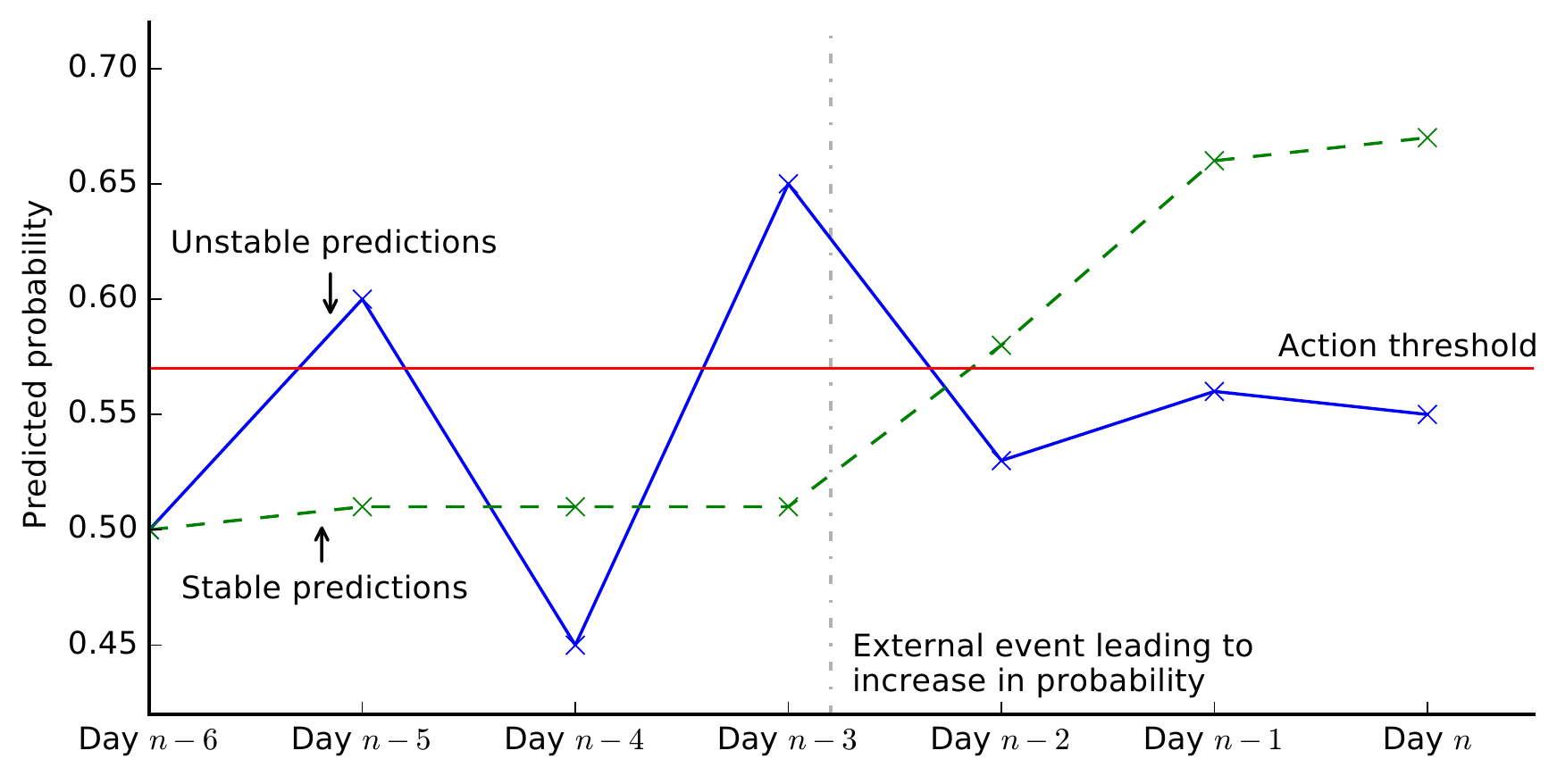}
    \end{center}
\caption{Illustration of the change in predicted probability on successive days, in a scenario where action is taken when the prediction is over a certain threshold (red horizontal line), and some external event leading to increase in probability occurred sometime between days $n-3$ and $n-2$ (indicated by the dot-dashed grey vertical line). The solid line (blue) and dashed line (green) shows the change in the predicted probability if the model does or does not produce a fluctuation in successive predictions respectively.}
\label{fig:risk_change_breakdown}
\end{figure}

% Prediction changes can be both endogenous, resulting from random model fluctuations and exogenous, due to a change in state of the system under measurement. 
% Some use of RF involves updating predictions over time and These applications require successive predictions from random forest to be \textit{stable}. 
% For applications involving tracking the predictions, ideally the predictions should be stable, as change = change due to actual activity (interested) + change due to fluctuation in prediction (uninterested)
% measuring prediction changes
It is often the case that changes in successive prediction values are more important than the absolute values. Examples include predicting disease risk \cite{Khalilia2011} and changes in customer lifetime value \cite{Chamberlain2017}. In these cases we wish to measure a change in the external environment. We call the variability in predictions due solely to changes in the external environment \textbf{exogenous variability}. Figure~\ref{fig:risk_change_breakdown} illustrates prediction changes with and without endogenous changes on top of exogenous change. Ideally we would like to measure only exogenous change, which is challenging if the endogenous effects are on a similar or larger scale. 

% Cost is also a concern, with Computation aaS emerging
Besides stability and error our framework also accounts for the cost of running the model. The emergence of computing as a service (Amazon elastic cloud, MS Azure etc.) makes the cost of running machine learning algorithms transparent and, for a given set of resources, proportional to runtime.

% Any hyper-parameter optimisation improves one or two in bias, stability, cost, but not all
It is not possible to find parameter configurations that simultaneously optimise cost, stability and error. For example, increasing the number of trees in a random forest will improve the stability of predictions, reduce the error, but increase the cost (due to longer runtimes). We propose a principled approach to this problem using a multi-criteria objective function.

We use Bayesian optimisation to search the parameter space of the multi-criteria objective function. Bayesian optimisation was originally developed by Kushner \cite{Kushner1964} and improved by Mockus \cite{Mockus1974}. It is a non-linear optimisation framework that has recently become popular in machine learning as it can find optimal parameter settings faster than competing methods such as random / grid search or gradient descent \cite{Snoek2012b}. The key idea is to perform a search over possible parameters that balances exploration (trying new regions of parameter space we know little about) with exploitation (choosing parts of the parameter space that are likely to lead to good objectives). This is achieved by placing a prior distribution on the mapping from parameters to the loss. An acquisition function then queries successive parameter settings by balancing high variance regions of the prior (good for exploration) with low mean regions (good for exploitation). The optimal parameter setting is then obtained as the setting with the lowest posterior mean after a predefined number of query iterations.

% experimental evaluation
We demonstrate the success of our approach on two large, public commercial datasets. 
% Contribution - a summary of our introduction
Our work makes the following contributions:
\begin{enumerate}
    \item A novel metric for the stability of the predictions of a model over different runs and its relationship with the variance and covariance of the predictions.
    \item A framework to optimise model hyperparameters and training parameters against the joint effect of prediction error, prediction stability and training cost, utilising constrained optimisation and Bayesian optimisation.
    \item A case study on the effects of changing hyperparameters of a random forest and training parameters on the model error, prediction stability and training cost, as applied on two publicly available datasets.
\end{enumerate}

The rest of the paper is organized as follows: in section \ref{sec:stability} we propose a novel metric to assess the stability of random forest predictions, in section \ref{sec:optimisation} we propose a random forest parameter tuning framework using a set of metrics, in section \ref{sec:hyperparameter_tuning} we discuss the effects of the hyper-parameters on the metrics and illustrate the usefulness of the proposed optimization framework to explore the trade-offs in the parameter space in section \ref{sec:experiments}.

\section{Prediction Stability}
\label{sec:stability}

Here we formalise the notion of random forest stability in terms of repeated model runs using the same parameter settings and dataset (ie. all variability is endogenous). The expected squared difference between the predictions over two runs is given by
\begin{align}
\frac{1}{N} \sum_{i=1}^{N} \left[\left(\hat{y}_i^{(j)} - \hat{y}_i^{(k)}\right)^2\right] ,
\end{align}
where $\hat{y}_i^{(j)} \in [0,1]$ is the probability from the $j^{th}$ run that the $i^{th}$ data point is of the positive class in binary classification problems (note this can be extended to multiclass classification and regression problems).  We average over $R \gg 1$ runs to give the Mean Squared Prediction Delta (MSPD):
\begin{align}
\textrm{MSPD}(f) &= \frac{2}{R(R-1)} \sum_{j=1}^R \sum_{k=1}^{j-1} \Bigg[ \frac{1}{N} \sum_{i=1}^{N} \left[\left(\hat{y}_i^{(j)} - \hat{y}_i^{(k)}\right)^2\right] \Bigg] \label{eq:rmspd_def}\\
&=  \frac{2}{N} \sum_{i=1}^N \Bigg[ \frac{1}{R-1} \sum_{l=1}^{R} \left(\hat{y}_i^{(l)} - \mathbb{E}(\hat{y}_i^{(.)})\right)^2 \notag\\
& \qquad\qquad\quad - \frac{1}{R(R-1)} \sum_{j=1}^{R} \sum_{k=1}^{R} \left(\hat{y}_i^{(j)} - \mathbb{E}(\hat{y}_i^{(.)})\right) \left(\hat{y}_i^{(k)} - \mathbb{E}(\hat{y}_i^{(.)})\right) \Bigg] \notag\\
&= 2\mathbb{E}_{x_i} [ \textrm{Var}(f(x_i)) - \textrm{Cov}(f_{j}(x_i),f_{k}(x_i))],
\label{eq:rmspd_def_two}
\end{align}

where $\mathbb{E}_{x_i}$ is the expectation over all validation data, $f$ is a mapping from a sample $x_i$ to a label $y_i$ on a given run, $\textrm{Var}(f(x_i))$ is the variance of the predictions of a single data point over model runs, and $ \textrm{Cov}(f_{j}(x_i),f_{k}(x_i))$ is the covariance of predictions of a single data point over two model runs.\footnote{A full derivation is available at our GitHub repository \url{https://github.com/liuchbryan/generalised_forest_tuning}.}

The covariance, the variance and hence the model instability are closely related to the forest parameter settings, which we discuss in section~\ref{sec:hyperparameter_tuning}. It is convenient to measure stability on the same scale as the forest predictions and so in the experiments we report the $\textrm{RMSPD} = \sqrt{\textrm{MSPD}}$.

\section{Parameter Optimisation Framework}
\label{sec:optimisation}
% intro
In industrial applications, where ultimately machine learning is a tool for profit maximisation, optimising parameter settings based solely on error metrics is inadequate. Here we develop a generalised loss function that incorporates our stability metric in addition to prediction error and running costs. We use this loss with Bayesian optimisation to select parameter values. 

% We present an optimisation framework to leverage the three practical considerations -- error reduction, prediction stability, and training cost.
\subsection{Metrics}
Before composing the loss function we define the three components:

\paragraph{Stability}
We incorporate stability (defined in section~\ref{sec:stability}) in to the optimization framework with the use of the $\textrm{RMSPD}$.

\paragraph{Error reduction}
Many different error metrics are used with random forests. These include F1-score, accuracy, precision, recall and Area Under the receiver operating characteristics Curve (AUC) and all such metrics fit within our framework. In the remainder of the paper we use the AUC because for binary classification, most other metrics require the specification of a threshold probability. As random forests are not inherently calibrated, a threshold of 0.5 may not be appropriate and so using AUC simplifies the exposition \cite{Chamberlain2017}. 

\paragraph{Cost reduction}
It is increasingly common for machine learning models to be run on the cloud with computing resources paid for by the hour (e.g. Amazon Web Services). Due to the exponential growth in data availability, the cost to run a model can be comparable with the financial benefit it produces. We use the training time (in seconds) as a proxy of the training cost.

\subsection{Loss-function}
We choose a loss function that is linear in cost, stability and AUC that allows the relative importance of these three considerations to be balanced:
\begin{align}
   L = \beta \,\textrm{RMSPD}(N_t, d, p) +
                      \gamma \,\textrm{Runtime}(N_t, d, p) - \alpha \,\textrm{AUC}(N_t, d, p)    ,               
\end{align}

where $N_t$ is the number of trees in the trained random forest, $d$ is the maximum depth of the trees, and $p$ is the proportion of data points used in training; $\alpha, \beta, \gamma$ are weight parameters. We restrict our analysis to three parameters of the random forest, but it can be easily extended to include additional parameters (e.g. number of features bootstrapped in each tree).

The weight parameters $\alpha, \beta$ and $\gamma$ are specified according to business/research needs. We recognise the diverse needs across different organisations and thus refrain from specifying what constitutes a ``good'' weight parameter set. Nonetheless, a way to obtain the weight parameters is to quantify the gain in AUC, the loss in RMSPD, and the time saved all in monetary units. For example, if calculations reveal 1\% gain in AUC equates to \pounds 50 potential business profit, 1\% loss in RMSPD equates to \pounds 10 reduction in lost business revenue, and a second of computation costs \pounds 0.01, then $\alpha, \beta$ and $\gamma$ can be set as 5,000, 1,000 and 0.01 respectively.

\subsection{Bayesian Optimisation}

The loss function is minimized using Bayesian optimisation. The use of Bayesian optimisation is motivated by the expensive, black-box nature of the objective function: each evaluation involves training multiple random forests, a complex process with internal workings that are usually masked from users. This rules out gradient ascent methods due to unavailability of derivatives. Exhaustive search strategies, such as grid search or random search, have prohibitive runtimes due to the large random forest parameter space.

A high-level overview on Bayesian Optimisation is provided in section \ref{sec:intro}.
Many different prior functions can be chosen and we use the Student-t process implemented in \texttt{pybo} \cite{Martinez-Cantin2014,Hoffman2014}.

\section{Parameter Sensitivity}
\label{sec:hyperparameter_tuning}

Here we describe three important random forest parameters and evaluate the sensitivity of our loss function to them.

\subsection{Sampling training data}

Sampling of training data -- drawing a random sample from the pool of available training data for model training -- is commonly employed to keep the training cost low. A reduction in the size of training data leads to shorter training times and thus reduces costs. However, reducing the amount of training data reduces the generalisability of the model as the estimator sees less training examples, leading to a reduction in AUC. Decreasing the training sample size also decreases the stability of the prediction. This can be understood by considering the form of the stability measure of $f$, the RMSPD (equation \ref{eq:rmspd_def}). The second term in this equation is the expected covariance of the predictions over multiple training runs. Increasing the size of the random sample drawn as training data increases the probability that the same input datum will be selected for multiple training runs and thus the covariance of the predictions increases. An increase in covariance leads to a reduction in the RMSPD (see equation \ref{eq:rmspd_def_two}).

% Improve stability: consider a tree as a random sample from a distribution with mean mu and variance sigma^2, a random forest is essentially taking the sample average of the distribution. By central limit theorem, the sample mean's (random forest) variance will drop due to having more sample (trees), which (since RMSPD is bounded by a multiple of variance) will decrease RMSPD.
% Improves AUC
% Cost a lot, training linear to number of trees
\subsection{Number of trees in a random forest}
Increasing the number of trees in a random forest will decrease the RMSPD (and hence improve stability) due to the Central Limit Theorem (CLT). Consider a tree in a random forest with training data bootstrapped. Its prediction can be seen as a random sample from a distribution with finite mean and variance~$\sigma^2$.\footnote{This could be any distribution as long as its first two moments are finite, which is usually the case in practice as predictions are normally bounded.} By averaging the trees' predictions, the random forest is computing the sample mean of the distribution. By the CLT, the sample mean will converge to a Gaussian distribution with variance~$\frac{\sigma^2}{N_t}$, where $N_t$ is the number of trees in the random forest. 

To link the variance to the MSPD, recall from equation \ref{eq:rmspd_def} that MSPD captures the interaction between the variance of the model and covariance of predictions between different runs:
\begin{align*}
\textrm{MSPD}(f) = 2\mathbb{E}_{x_i} [ \textrm{Var}(f(x_i)) - \textrm{Cov}(f_{j}(x_i),f_{k}(x_i))].
\end{align*}

The covariance is bounded below by the negative square root of the variance of its two elements, which is in turn bounded below by the negative square root of the larger variance squared:
\begin{align}
   \textrm{Cov}(f_{j}(x_i),f_{k}(x_i)) 
\geq &-\sqrt{\textrm{Var}(f_{j}(x_i)) \textrm{Var}(f_{k}(x_i))} \notag\\
\geq &-\sqrt{(\max\{\textrm{Var}(f_{j}(x_i)), \textrm{Var}(f_{k}(x_i))\})^2}.
\label{eq:cov}
\end{align}

Given $f_j$ and $f_k$ have the same variance as $f$ (being the models with the same training proportion across different runs), the inequality~\ref{eq:cov} can be simplified as:
\begin{align}
   \textrm{Cov}(f_{j}(x_i),f_{k}(x_i)) 
\geq &-\sqrt{(\max\{\textrm{Var}(f(x_i)), \textrm{Var}(f(x_i))\})^2} = -\textrm{Var}(f(x_i)).
\label{eq:cov2}
\end{align}

MSPD is then bounded above by a multiple of the expected variance of $f$:
\begin{align}
   \textrm{MSPD}(f) \leq 2\mathbb{E}_{x_i} [ \textrm{Var}(f(x_i)) - (-\textrm{Var}(f(x_i)))] = 4\mathbb{E}_{x_i} [ \textrm{Var}(f(x_i))],
 \label{eq:mspd_ineq}
\end{align}
which decreases as $N_t$ increases, leading to a lower RMSPD estimate.

While increasing the number of trees in a random forest reduces error and improves stability in predictions, it increases the training time and hence monetary cost. In general, the runtime complexity for training a random forest grows linearly with the number of trees in the forest.

\subsection{Maximum depth of a tree}
The maximum tree depth controls the complexity of each decision tree and the computational cost (running time) increases exponentially with tree depth. The optimal depth for error reduction depends on the other forest paramaters and the data. Too much depth causes overfitting. Additionally, as the depth increases the prediction stability will decrease as each model tends towards memorizing the training data. The highest stability will be attained using shallow trees, however if the forest is too shallow the model will underfit resulting in low AUC.

\section{Experiments}
\label{sec:experiments}

We evaluate our methodology by performing experiments on two public  datasets: (1) the Orange small dataset from the 2009 KDD Cup and (2) the Criteo display advertising challenge Kaggle competition from 2014. Both datasets have a mixture of numerical and categorical features and binary target labels (Orange: 190 numerical, 40 categorical, Criteo: 12 numerical, 25 categorical). 

We report the results of two sets of experiments: (1) Evaluating the effect of changing random forest parameters on the stability and loss functions (2) Bayesian optimisation with different weight parameters.

We train random forests to predict the upselling label for the Orange dataset and the click-through rate for the Criteo dataset. Basic pre-processing steps were performed on both datasets to standardise the numerical data and transform categoricals into binary indicator variables. We split the datasets into two halves: the first as training data (which may be further sampled at each training run), and the later as validation data. All data and code required to replicate our experiments is available from our GitHub repository.\footnote{\url{https://github.com/liuchbryan/generalised_forest_tuning}}

\subsection{Parameter Sensitivity}

In the first set of experiments we evaluate the effect of varying random forest parameters on the components of our loss function.

\begin{figure}
    \begin{center}   \includegraphics[width=\textwidth]{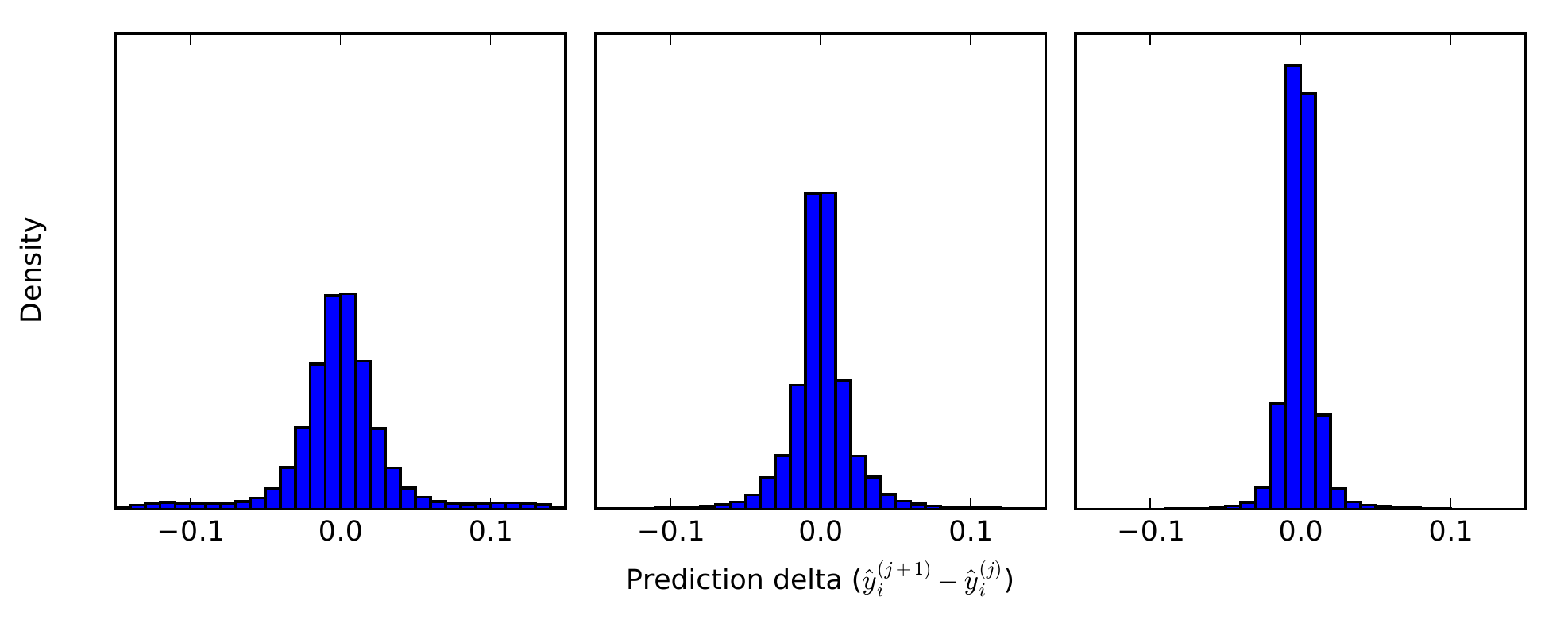}
    \end{center}
\caption{The distribution of prediction deltas (difference between two predictions on the same validation datum) for successive runs of random forests with (from left to right) 8, 32, and 128 trees, repeated ten times. The RMSPD for these three random forests are 0.046, 0.025, and 0.012 respectively. Training and prediction are done on the Orange small dataset with upselling labels. The dataset is split into two halves: the first 25k rows are used for training the random forests, and the latter 25k rows for making predictions. Each run re-trains on all 25k training data, with trees limited to a maximum depth of 10.}
\label{fig:prediction_delta_distribution}
\end{figure}

Figure~\ref{fig:prediction_delta_distribution} visualises the change in the RMSPD with relation to the number of trees in the random forest. The plots show distributions of prediction deltas for the Orange dataset. Increasing the number of trees (going from the left to the right plot) leads to a more concentrated prediction delta distribution, a quality also reflected by a reduction in the RMSPD.

\begin{figure}
    \begin{center}   
    \includegraphics[width=0.328\textwidth]{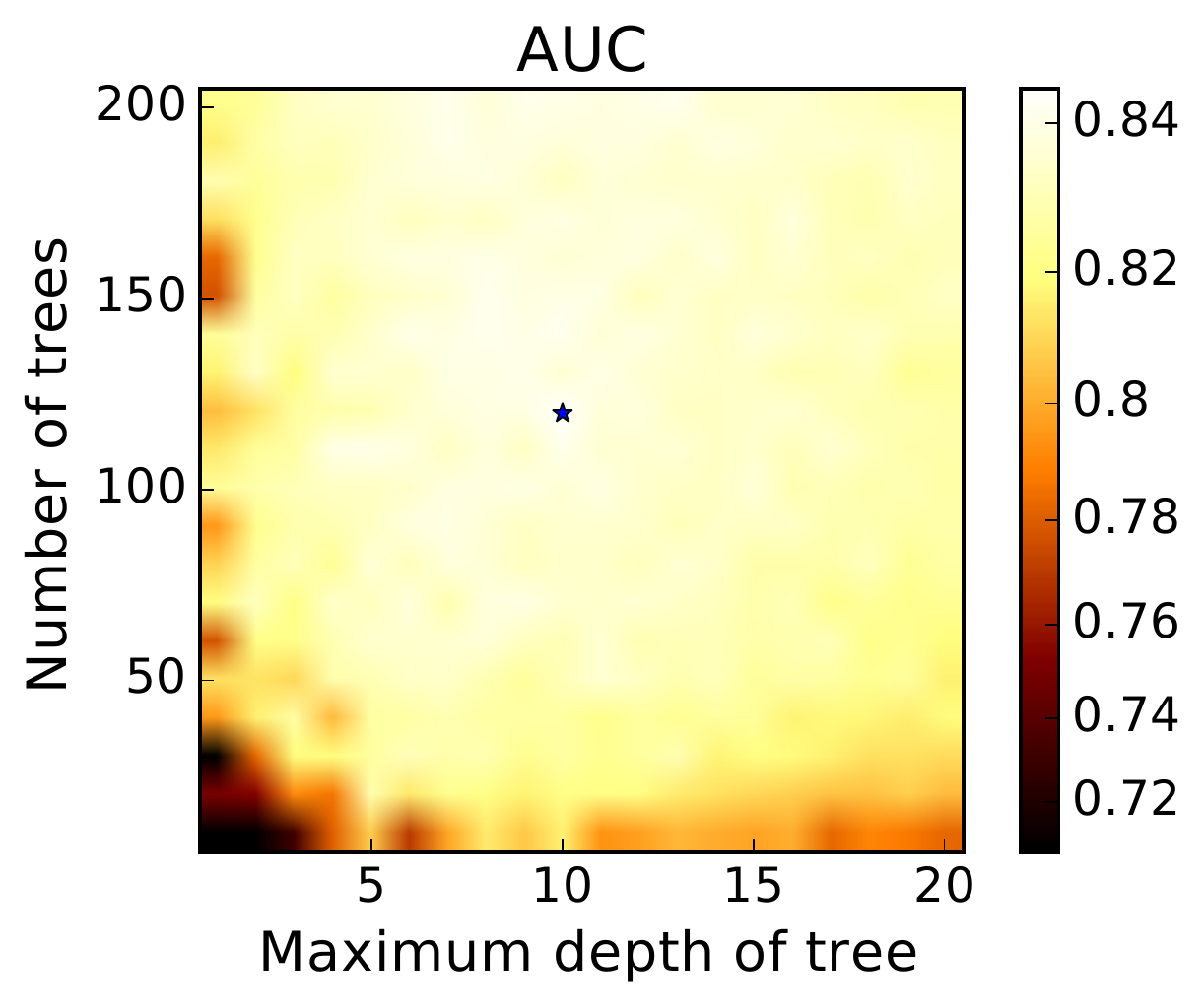}
    \includegraphics[width=0.335\textwidth]{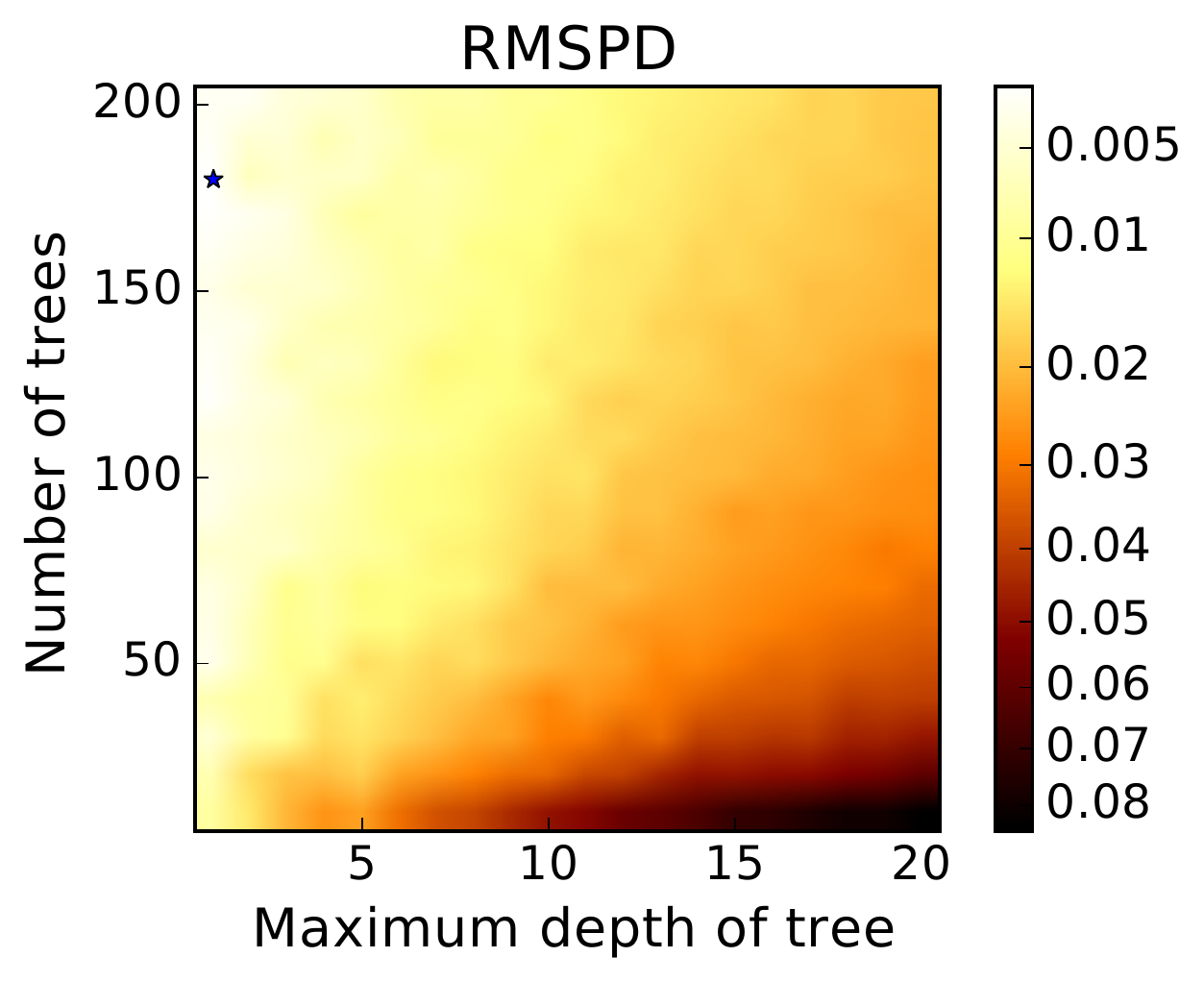}
    \includegraphics[width=0.317\textwidth]{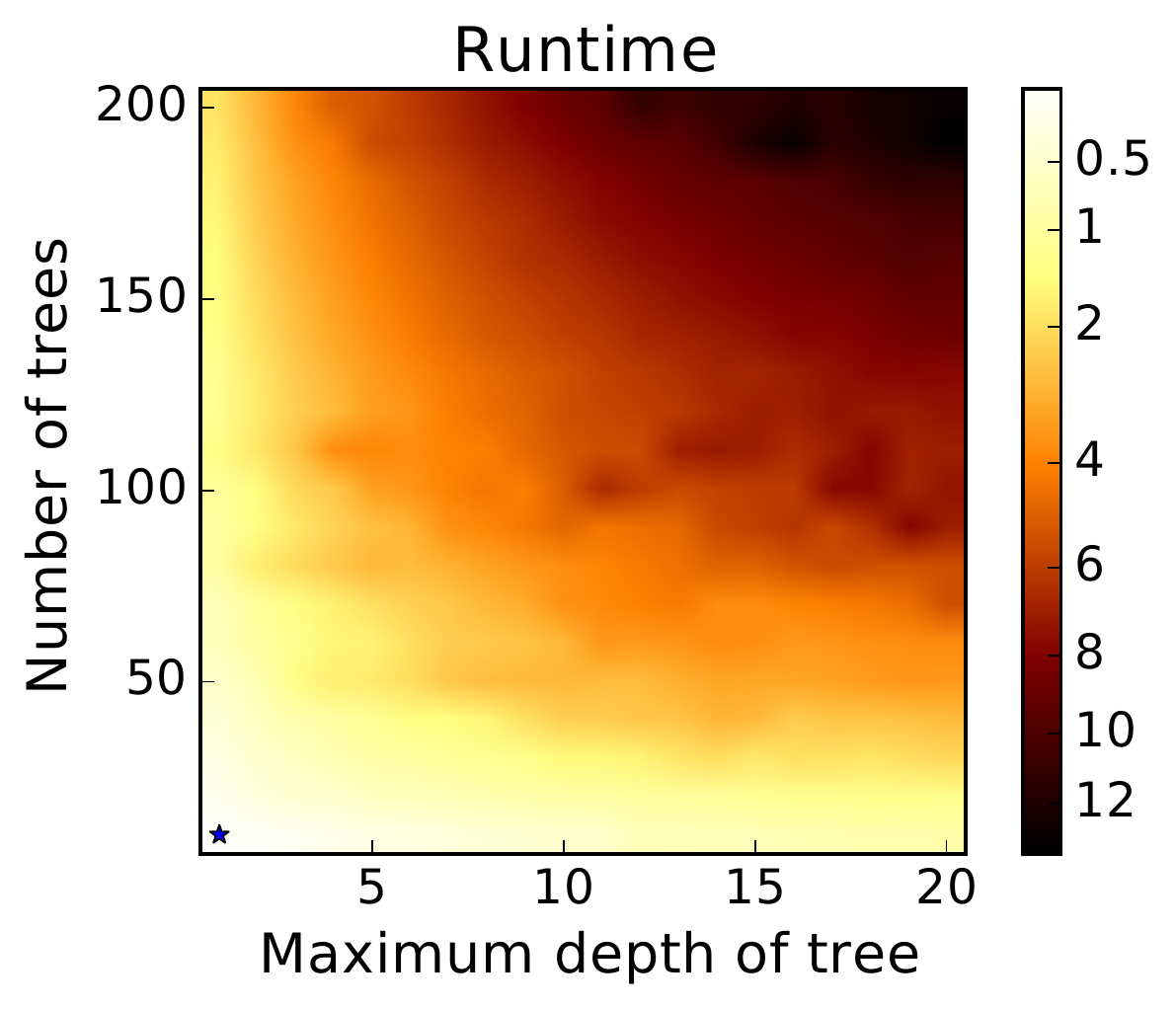} \\ \vspace{3mm}
        \includegraphics[width=0.37\textwidth]{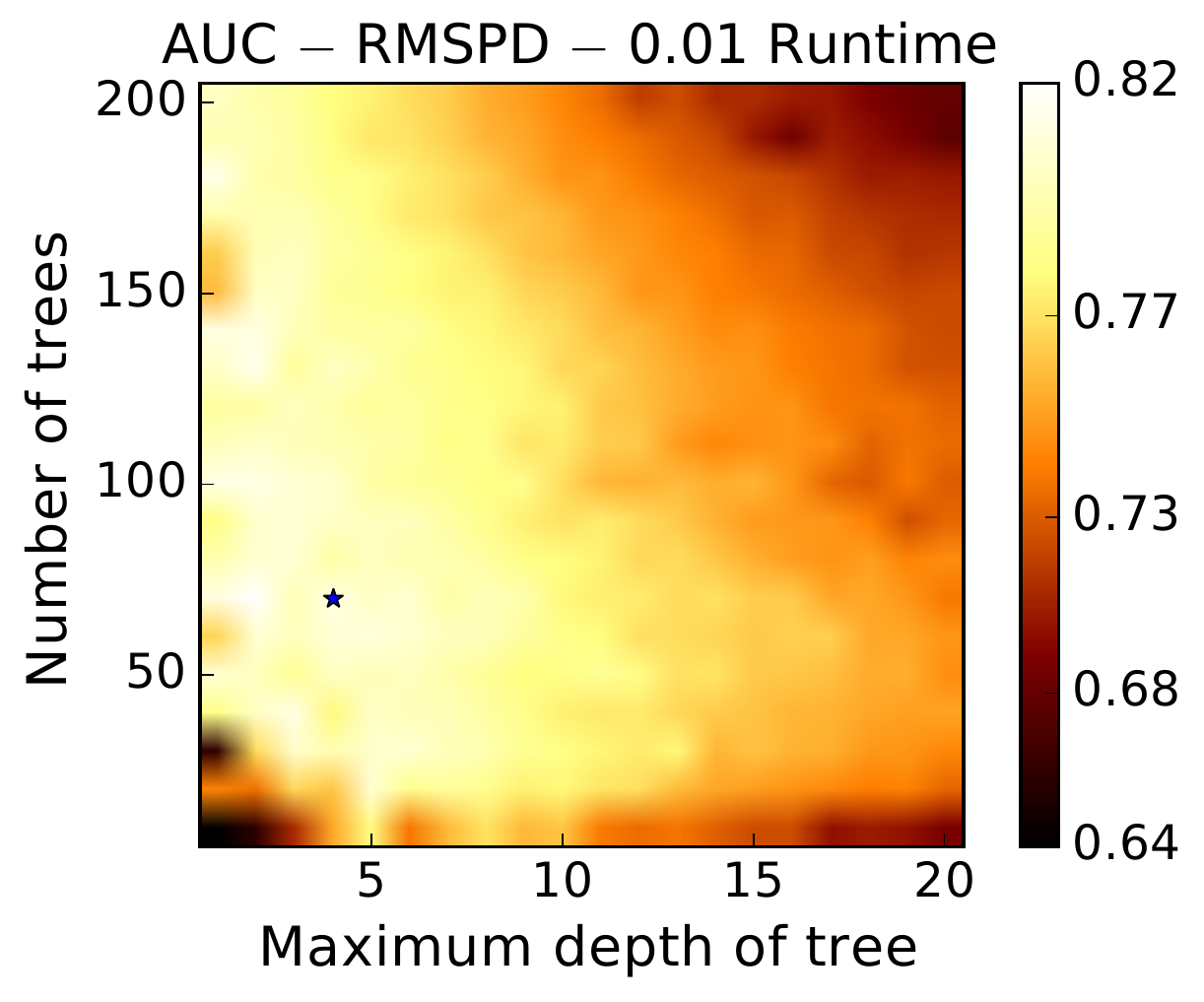}
    \includegraphics[width=0.37\textwidth]{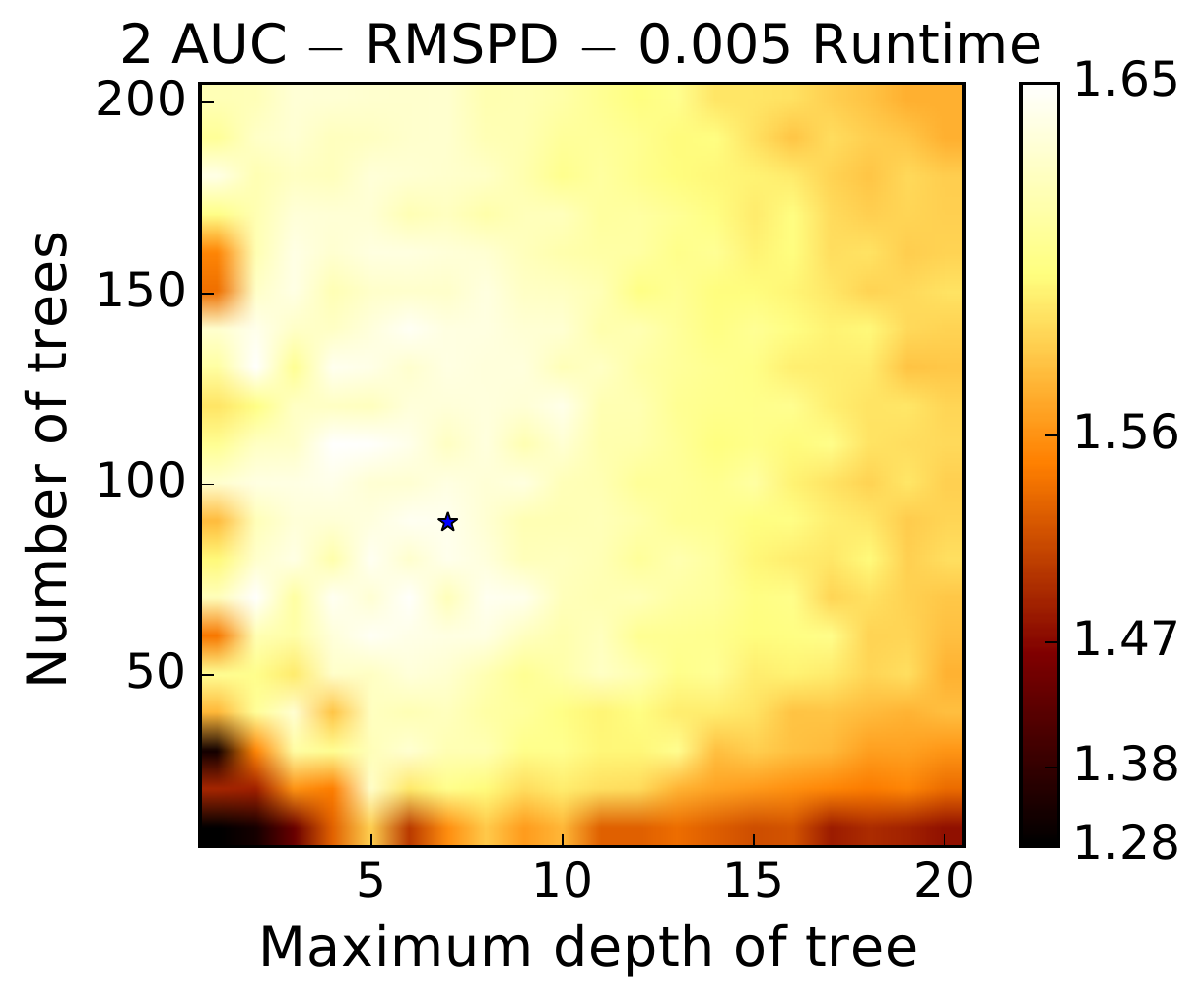}
    \end{center}
\caption{The average AUC (top left), RMSPD (top middle), and average runtime (top right) attained by random forests with different number of trees and maximum tree depth (training proportion is fixed at 0.5) over five train/test runs, as applied on the Orange dataset. The bottom two plots shows the value attained in the specified objective functions by the random forests above.
A lighter spot on the maps represents a more preferable parametrization. The shading is scaled between the minimum and maximum values in each chart. The optimal configuration found under each metric is indicated by a blue star.}
\label{fig:trees_depth}
\end{figure}

Figure \ref{fig:trees_depth} shows the AUC, runtime, RMSPD and loss functions averaged over multiple runs of the forest for different settings of number of trees and maximum tree depth. It shows that the AUC plateaus for a wide range of combinations of number of trees and maximum depth. The RMSPD is optimal for large numbers of shallow trees while runtime is optimised by few shallow trees. When we form a linear combination of the three metrics, the optimal solutions are markedly different from those discovered by optimising any single metric in isolation. We show this for $\alpha = 1, \beta = 1, \gamma = 0.01$ and $\alpha = 2, \beta = 1, \gamma = 0.005$.  

\subsection{Bayesian optimization of the trilemma}
We also report the results of using the framework to choose the parameters.
The aim of these experiments is to show that (1) Bayesian optimisation provides a set of parameters that achieve good AUC, RMSPD and runtime, and (2) by varying the weight parameters in the Bayesian optimisation a user is able to prioritise one or two of the three respective items. 

Table \ref{tab:hyperparameter_tuning_summary} summarises the trilemma we are facing -- all three parameter tuning strategies improves two of the three practical considerations with the expense of the consideration(s) left. 

\begin{table}[!ht]
  \begin{center}
  \begin{tabularx}{0.95\textwidth}{p{0.4\textwidth} || X | X | X}
     Hyperparameter tuning strategy & AUC gain & RMSPD reduction & Cost savings \\ \hline
     Increase training proportion & + & + & -- \\
     Increase number of trees & + & + & -- -- \\
     Reduce maximum depth of trees & -- & + & ++ \\

  \end{tabularx}
  \end{center}
  \caption{Effect of the common hyperparameter tuning strategies on the three practical considerations. Plus sign(s) means a positive effect to the measure (and hence more preferred), and minus sign(s) means a negative effect to the measure (and hence not preferred). The more plus/minus sign within the entry, the more prominent the effect of the corresponding strategy.}
  \label{tab:hyperparameter_tuning_summary}
\end{table}

The results of our experiments on Bayesian optimisation of the trilemma are shown in tables \ref{tab:orange_bayes_opt} and \ref{tab:criteo_bayes_opt}. The first row in both tables shows the results for a vanilla random forest with no optimisation of the hyper-parameters discussed in the previous section: 10 trees, no limit on the maximum depth of the tree, and using the entire training data set (no sampling). The Bayesian optimisation for each set of weight parameters was run for 20 iterations, with the RMSPD calculated over three training runs in each iteration. 

The first observation from both sets of results is that Bayesian optimisation is suitable for providing a user with a framework that can simultaneously improve AUC, RMSPD and runtime as compared to the baseline. Secondly, it is clear that by varying the weight parameters, Bayesian optimisation is also capable of prioritising specifically AUC, RMSPD or runtime. Take for example the third and fourth rows of table \ref{tab:orange_bayes_opt}; setting $\beta = 5$ we see a significant reduction in the RMSPD in comparison to the second row where $\beta = 1$. Similarly, comparing the fourth row to the second row, increasing $\alpha$ from 1 to 5 gives a 1\% increase in AUC. In the final row we see that optimising for a short runtime keeps the RMSPD low in comparison to the non-optimal results on the first row and sacrifices the AUC instead.

\bgroup
\def\arraystretch{1.2}
\begin{table}[!ht]
\centering

\begin{tabular}{ccccccccccccccccc}
\hline
$\alpha$ && $\beta$ && $\gamma$ && $N_t^{*}$ && $d^{*}$ && $p^{*}$ && AUC && RMSPD && Runtime  \\
\hline
\hline
\multicolumn{11}{c}{No optimisation:} && 0.760  &&	0.112  &&	1.572 \\
\hline
1 &&	1	&& 0.01 &&	166 && 6 && 0.100&&	0.829 &&	0.011 &&	1.142 \\
1	&& \bf{5}	&& 0.01	 && 174 &&	1 && 	0.538 &&	0.829 &&	\bf{0.002} &&	1.452 \\
\bf{5} &&	1 &&	0.01 &&	144 &&	12 &&	0.583 &&	\bf{0.839} &&	0.013 &&	5.292\\
1	&& 1	 && \bf{0.05} && 158 &&	4	&& 0.100 &&	0.8315 && 0.0082 &&	\bf{1.029}\\
\hline
\end{tabular}
\vspace{0.4cm}
\caption{Results of Bayesian optimisation for the Orange dataset at various settings of $\alpha$, $\beta$ and $\gamma$, the weight parameters for the AUC, RMSPD and runtime respectively. The Bayesian optimiser has the ability to tune three random forest hyper-parameters: the number of trees, $N_t^{*}$, the maximum tree depth, $d^{*}$, and size of the training sample, $p^{*}$. Key results are emboldened and discussed further in the text.}
\label{tab:orange_bayes_opt}
\end{table}
\egroup

For the Criteo dataset (table \ref{tab:criteo_bayes_opt}) we see on the second and third row that again increasing the $\beta$ parameter leads to a large reduction in the RMSPD. For this dataset the Bayesian optimiser is more reluctant to use a larger number of estimators to increase AUC because the Criteo dataset is significantly larger (around 100 times) than the Orange dataset and so using more trees increases the runtime more severely. To force the optimiser to use more estimators we reduce the priority of the runtime by a factor of ten as can be seen in the final two rows. We see in the final row that doubling the importance of the AUC ($\alpha$) leads to a significant increase in AUC ($4.5\%$) when compared to the non-optimal results.

\bgroup
\def\arraystretch{1.2}
\begin{table}[!ht]
\vspace{-5mm}
\centering
\begin{tabular}{ccccccccccccccccc}
\hline
$\alpha$ && $\beta$ && $\gamma$ && $N_t^{*}$ && $d^{*}$ && $p^{*}$ && AUC && RMSPD && Runtime  \\
\hline
\hline
\multicolumn{11}{c}{No optimisation:} && 0.685  &&	0.1814  &&	56.196 \\
\hline
1  &&	\bf{1}  &&	0.01  &&	6  &&	8  &&	0.1  &&	0.7076  &&	\bf{0.04673}  &&	1.897 \\
1  &&	\bf{5}  &&	0.01  &&	63  &&	3  &&	0.1  &&	0.6936 &&	\bf{0.01081} &&	4.495 \\
1 &&	1 &&	0.05 &&	5 &&	5 &&	0.1 &&	0.688 &&	0.045 &&	1.136 \\
2  &&	1  &&	0.05  &&	9  &&	9  &&	0.1  &&	0.7145  &&	0.03843  &&	2.551  \\

1 &&	1  &&	0.001 &&	120 &&	2 &&	0.1 &&	0.6897 &&	0.007481 &&	7.153 \\
\bf{2} &&	1 &&	0.001 &&	66 &&	15 &&	0.1 &&	\bf{0.7300} &&	0.02059 &&	11.633 \\
\hline
\end{tabular}
\vspace{0.4cm}
\caption{Results of Bayesian optimisation for the Criteo dataset. The table shows the results of the Bayesian optimisation by varying $\alpha$, $\beta$ and $\gamma$ which control the importance of the AUC, RMSPD and runtime respectively. The Bayesian optimiser has the ability to tune three hyper-parameters of the the random forest: the number of trees, $N_t^{*}$, the maximum depth of the tree, $d^{*}$, and size of the training sample, $p^{*}$. Key results are emboldened and discussed further in the text.}
\vspace{-5mm}
\label{tab:criteo_bayes_opt}
\end{table}
\egroup

\section{Conclusion}
\label{sec:conclusion}
We proposed a novel metric to capture the stability of random forest predictions, which is key for applications where random forest models are continuously updated. We show how this metric, calculated on a sample, is related to the variance and covariance of the predictions over different runs. While we focused on random forests in this text, the proposed stability metric is generic and can be applied to other non-deterministic models (e.g. gradient boosted trees, deep neural networks) as well as deterministic training methods when training is done with a subset of the available data. 

We also propose a framework for multi-criteria optimisation, using the proposed metric in addition to metrics measuring error and cost. We validate this approach using two public datasets and show how optimising a model solely for error can lead to poorly specified parameters. 

%
% ---- Bibliography ----
%

\bibliographystyle{splncs03}
% \bibliography{Mendeley_ASOS}
% \bibliography{references.bbl}

\appendix

\end{document}